\documentclass[conference]{IEEEtran}
\IEEEoverridecommandlockouts
\usepackage{mathtools}
\usepackage{tikz}
\usetikzlibrary{tikzmark}

\usepackage{lipsum}
\usepackage{amssymb,mathtools}
\usepackage{algorithmic}
\usepackage{graphicx}
\usepackage[colorinlistoftodos, textwidth=3cm]{todonotes} 
\usepackage{tikz}
\usetikzlibrary{positioning}
\usepackage{textcomp}
\usepackage{xcolor}
\def\BibTeX{{\rm B\kern-.05em{\sc i\kern-.025em b}\kern-.08em
    T\kern-.1667em\lower.7ex\hbox{E}\kern-.125emX}}

\begin{document}

\title{Untangling Braids with Multi-agent Q-Learning\\
{\footnotesize \textsuperscript{}}
\thanks{This work was supported by the Leverhulme Trust Research Project Grant RPG-2019-313.}
}

\author{\IEEEauthorblockN{\textsuperscript{} Abdullah Khan}
\IEEEauthorblockA{\textit{Department of Mathematical Sciences} \\
\textit{University of Essex}\\
Colchester, UK\\
ak20749@essex.ac.uk}
\and
\IEEEauthorblockN{\textsuperscript{} Alexei Vernitski}
\IEEEauthorblockA{\textit{Department of Mathematical Sciences} \\
\textit{University of Essex}\\
Colchester, UK \\
asvern@essex.ac.uk}
\and
\IEEEauthorblockN{\textsuperscript{} Alexei Lisitsa}
\IEEEauthorblockA{\textit{Department of Computer Science} \\
\textit{University of Liverpool}\\
Liverpool, UK\\
A.Lisitsa@liverpool.ac.uk}

}

\maketitle

\begin{abstract}
We use reinforcement learning to tackle the problem of untangling braids. We experiment with braids with 2 and 3 strands. Two competing players learn to tangle and untangle a braid. We interface the braid untangling problem with the OpenAI Gym environment, a widely used way of connecting agents to reinforcement learning problems. The results provide evidence that the more we train the system, the better the untangling player gets at untangling  braids. At the same time, our tangling player produces good examples of tangled braids. 

\end{abstract}


\section[Introduction]{Introduction}
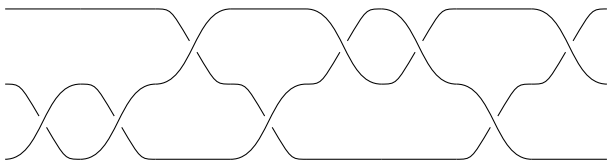
\begin{figure}[h]
\centering
\begin{tikzpicture}[scale=0.50]
\draw (0,0) .. controls (1,0) and (1,2) .. (2,2);
\draw (0,2) .. controls (0.4,2)  .. (0.9,1.15);
\draw (2,0) .. controls (1.6,0)  .. (1.1,0.85);
\draw (0,4) -- (2,4);
\draw (2,0) .. controls (3,0) and (3,2) .. (4,2);
\draw (2,2) .. controls (2.4,2)  .. (2.9,1.15);
\draw (4,0) .. controls (3.6,0)  .. (3.1,0.85);
\draw (2,4) -- (4,4);
\draw (4,0) -- (6,0);
\draw (4,2) .. controls (5,2) and (5,4) .. (6,4);
\draw (4,4) .. controls (4.4,4)  .. (4.9,3.15);
\draw (6,2) .. controls (5.6,2)  .. (5.1,2.85);
\draw (6,0) .. controls (7,0) and (7,2) .. (8,2);
\draw (6,2) .. controls (6.4,2)  .. (6.9,1.15);
\draw (8,0) .. controls (7.6,0)  .. (7.1,0.85);
\draw (6,4) -- (8,4);
\draw (8,0) -- (10,0);
\draw (8,4) .. controls (9,4) and (9,2) .. (10,2);
\draw (8,2) .. controls (8.4,2)  .. (8.9,2.85);
\draw (10,4) .. controls (9.6,4)  .. (9.1,3.15);
\draw (10,0) -- (12,0);
\draw (10,4) .. controls (11,4) and (11,2) .. (12,2);
\draw (10,2) .. controls (10.4,2)  .. (10.9,2.85);
\draw (12,4) .. controls (11.6,4)  .. (11.1,3.15);
\draw (12,2) .. controls (13,2) and (13,0) .. (14,0);
\draw (12,0) .. controls (12.4,0)  .. (12.9,0.85);
\draw (14,2) .. controls (13.6,2)  .. (13.1,1.15);
\draw (12,4) -- (14,4);
\draw (14,0) -- (16,0);
\draw (14,4) .. controls (15,4) and (15,2) .. (16,2);
\draw (14,2) .. controls (14.4,2)  .. (14.9,2.85);
\draw (16,4) .. controls (15.6,4)  .. (15.1,3.15);
\end{tikzpicture}
\caption{Braid $aabaBBAB$}  \label{braid1}
\end{figure}

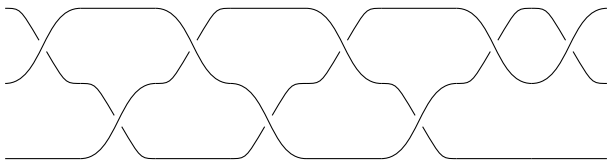
\begin{figure}[h]
\centering
\begin{tikzpicture}[scale=0.50]
\draw (0,0) -- (2,0);
\draw (0,2) .. controls (1,2) and (1,4) .. (2,4);
\draw (0,4) .. controls (0.4,4)  .. (0.9,3.15);
\draw (2,2) .. controls (1.6,2)  .. (1.1,2.85);
\draw (2,0) .. controls (3,0) and (3,2) .. (4,2);
\draw (2,2) .. controls (2.4,2)  .. (2.9,1.15);
\draw (4,0) .. controls (3.6,0)  .. (3.1,0.85);
\draw (2,4) -- (4,4);
\draw (4,0) -- (6,0);
\draw (4,4) .. controls (5,4) and (5,2) .. (6,2);
\draw (4,2) .. controls (4.4,2)  .. (4.9,2.85);
\draw (6,4) .. controls (5.6,4)  .. (5.1,3.15);
\draw (6,2) .. controls (7,2) and (7,0) .. (8,0);
\draw (6,0) .. controls (6.4,0)  .. (6.9,0.85);
\draw (8,2) .. controls (7.6,2)  .. (7.1,1.15);
\draw (6,4) -- (8,4);
\draw (8,0) -- (10,0);
\draw (8,4) .. controls (9,4) and (9,2) .. (10,2);
\draw (8,2) .. controls (8.4,2)  .. (8.9,2.85);
\draw (10,4) .. controls (9.6,4)  .. (9.1,3.15);
\draw (10,0) .. controls (11,0) and (11,2) .. (12,2);
\draw (10,2) .. controls (10.4,2)  .. (10.9,1.15);
\draw (12,0) .. controls (11.6,0)  .. (11.1,0.85);
\draw (10,4) -- (12,4);
\draw (12,0) -- (14,0);
\draw (12,4) .. controls (13,4) and (13,2) .. (14,2);
\draw (12,2) .. controls (12.4,2)  .. (12.9,2.85);
\draw (14,4) .. controls (13.6,4)  .. (13.1,3.15);
\draw (14,0) -- (16,0);
\draw (14,2) .. controls (15,2) and (15,4) .. (16,4);
\draw (14,4) .. controls (14.4,4)  .. (14.9,3.15);
\draw (16,2) .. controls (15.6,2)  .. (15.1,2.85);
\end{tikzpicture}
\caption{Braid $baBABaBb$} \label{braid2}
\end{figure}

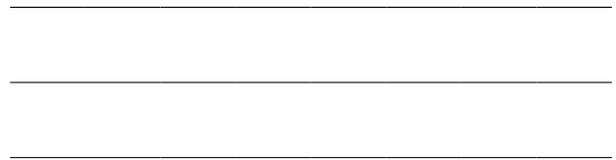
\begin{figure}[h]
\centering
\begin{tikzpicture}[scale=0.50]
\draw (0,0) -- (2,0);
\draw (0,2) -- (2,2);
\draw (0,4) -- (2,4);
\draw (2,0) -- (4,0);
\draw (2,2) -- (4,2);
\draw (2,4) -- (4,4);
\draw (4,0) -- (6,0);
\draw (4,2) -- (6,2);
\draw (4,4) -- (6,4);
\draw (6,0) -- (8,0);
\draw (6,2) -- (8,2);
\draw (6,4) -- (8,4);
\draw (8,0) -- (10,0);
\draw (8,2) -- (10,2);
\draw (8,4) -- (10,4);
\draw (10,0) -- (12,0);
\draw (10,2) -- (12,2);
\draw (10,4) -- (12,4);
\draw (12,0) -- (14,0);
\draw (12,2) -- (14,2);
\draw (12,4) -- (14,4);
\draw (14,0) -- (16,0);
\draw (14,2) -- (16,2);
\draw (14,4) -- (16,4);
\end{tikzpicture}
\caption{The trivial braid without intersections}  \label{braid-trivial}
\end{figure}

Braids are mathematical objects from low-dimensional topology which can be successfully encoded with sequences of letters and, therefore, studied using algebra or, as we do in this study, using some computer-scientific approach. 
 A braid on $n$ strands consists of $n$ ropes whose left-hand ends are fixed one under another and whose right-hand ends are fixed one under another; you can imagine that the braid is laid out on a table, and the ends of the ropes are attached to the table with nails. Figures \ref{braid1}, \ref{braid2}, \ref{braid-trivial} show examples of braids on $3$ strands. 

Two braids are \emph{equivalent} to one another if they can be transformed into one another by shifting and twisting the middle parts of the ropes (without touching the ends of the ropes). For example, the two braids in Figures \ref{braid1}, \ref{braid2} are equivalent to one another, although it is difficult to see it. They are also what is called \emph{trivial} braids, in the sense that they are equivalent to the braid without any intersections of ropes, shown in Figure \ref{braid-trivial}.

\begin{figure}
\centering
\begin{minipage}{.15\columnwidth}
  \centering
\begin{tikzpicture}[scale=0.5]
\draw (0,2) .. controls (1,2) and (1,0) .. (2,0);
\draw (0,0) .. controls (0.4,0)  .. (0.9,0.85);
\draw (2,2) .. controls (1.6,2)  .. (1.1,1.15);
\draw (0,4) -- (2,4);
\end{tikzpicture}
\end{minipage}%
\begin{minipage}{.15\columnwidth}
  \centering
\begin{tikzpicture}[scale=0.5]
\draw (0,0) .. controls (1,0) and (1,2) .. (2,2);
\draw (0,2) .. controls (0.4,2)  .. (0.9,1.15);
\draw (2,0) .. controls (1.6,0)  .. (1.1,0.85);
\draw (0,4) -- (2,4);
\end{tikzpicture}
\end{minipage}
\begin{minipage}{.15\columnwidth}
  \centering
\begin{tikzpicture}[scale=0.5]
\draw (0,0) -- (2,0);
\draw (0,4) .. controls (1,4) and (1,2) .. (2,2);
\draw (0,2) .. controls (0.4,2)  .. (0.9,2.85);
\draw (2,4) .. controls (1.6,4)  .. (1.1,3.15);
\end{tikzpicture}
\end{minipage}
\begin{minipage}{.15\columnwidth}
  \centering
\begin{tikzpicture}[scale=0.5]
\draw (0,0) -- (2,0);
\draw (0,2) .. controls (1,2) and (1,4) .. (2,4);
\draw (0,4) .. controls (0.4,4)  .. (0.9,3.15);
\draw (2,2) .. controls (1.6,2)  .. (1.1,2.85);
\end{tikzpicture}
\end{minipage}
\begin{minipage}{.15\columnwidth}
  \centering
\begin{tikzpicture}[scale=0.5]
\draw (0,0) -- (2,0);
\draw (0,2) -- (2,2);
\draw (0,4) -- (2,4);
\end{tikzpicture}
\end{minipage}
\caption{Braid fragments $A, a, B, b, 1$}  \label{braid-fragments}
\end{figure}
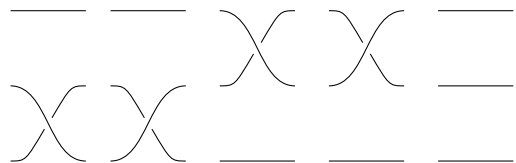

Now let us explain how braids can be represented conveniently in the computer. A braid is considered as a sequence of its simple fragments; for braids on $3$ strands, these are the fragments shown in Figure \ref{braid-fragments}, which we denote by $A, a, B, b, 1$ (and which in mathematical papers are usually denoted by $\sigma_1, \sigma_1^{-1}, \sigma_2, \sigma_2^{-1}, 1$).

Using this convenient notation, we can now say that the braids in Figures \ref{braid1}, \ref{braid2} are $aabaBBAB$ and $baBABaBb$. This notation is useful not only for describing braids, but also for checking if two braids are equivalent. Indeed, it is known that two braids are equivalent if and only if one can be transformed to the other using rules called the second Reidemeister move and the third Reidemeister move.
The \emph{second Reidemeister move} is the rule stating that $Aa$ and $aA$ are equivalent to $11$, and $Bb$ and $bB$ are also equivalent to $11$. (An algebraist studying braids in the context of group theory would also add that 11 is equivalent to 1; however, we felt that the performance of our AI will be best if we omit this non-essential rule.) The \emph{third Reidemeister move} is the rule stating that $ABA$ is equivalent to $BAB$.
Our general aim is to produce tangled braids and to untangle braids using reinforcement learning (RL). A recent study \cite{bib19} uses RL to untangle knots using a version of Reidemeister moves known as Markov moves. The novelty of our approach is the use two agents: one for tangling and one for untangling. 


In this pilot study we concentrate on braids with $2$ and $3$ strands. For braids with $2$ strands the problem is equivalent to simplifying words in a group given by a presentation $\langle a, b | ab=ba=1 \rangle$. In our experiments we choose to use the moves which preserve the length of the braid, for example, $ab$ simplifies to $11$ and not $1$. We approach the problem of untangling  braids with two strands as a symbol game. The input string of length n will consist of 3 symbols: ['$a$’, ‘$b$', ‘$1$’]. The task of the untangling agent is to convert the string to have all characters as ‘$1$’ (untangled state).
Following are the allowed moves:
$1a=a1$;      
$1b=b1$;
$ab=ba=11$. (In another experiment we used the moves
$1a=a1$;      
$1b=b1$;
$aa=bb=11$, corresponding to the group $\langle a, b | aa=bb=1 \rangle$.)
All such moves are implemented in both directions.

We also experimented with braids with three strands, where following transformations are allowed, Aa=aA=11; Bb=bB=11; A1=1A; a1=1a; B1=1B; b1=1b. We approach the problem of untangling braids on 3 strands as a game played between two players,  player 1 (the tangling player) and player 2 (the untangling player). Player 1 starts with an untangled braid as in Figure \ref{braid-trivial} and applies Reidemeister moves to tangle the braid. For example, braids in Figures \ref{braid1}, \ref{braid2} were produced by player 1 after approximately 150 games against player 2. Once player 1 has created a tangled braid after a fixed number of steps, that would be the input for the player 2 (untangling player); the task of player 2 is to apply Reidemeister moves to reach a fixed target output, that would be all 1's (untangled state).

In our experiment we approach the problem of untangling braids on $2$ and $3$ strands by simply using the Q-learning algorithm. Q-learning starting from the current state of the agent finds an optimal policy in the sense of maximizing the expected value of the total rewards \cite{bib18}. 
To implement the Q-learning algorithm, we use OpenAI Gym \cite{bib1}. It is an interface which provides a number of environments to implement reinforcement learning problems. The benefit of interfacing with OpenAI Gym is that it is an actively developed interface which allows to add environments and features useful while training the model. 

 
The  paper  is  organized  as  follows:  in  the  following section  we  discuss the Background taking into consideration basics of Reinforcement Learning with focus towards a technique known as Q-Learning. In Section 3, we briefly  review  the concept of OpenAI Gym, and how we have used it for our Problem. In Section 4, we mention about the experimental details and results. 
 
\section{Background}

In this section we formally highlight the important concepts for the understanding and development of the project, and also highlight some of the relevant work in the domain of reinforcement learning specifically for games.\newline 
Reinforcement learning is the training of machine learning models to make a sequence of decisions, where the agent learns to achieve a goal in an uncertain, potentially complex environment\cite{bib15}. In RL, there is a game-like situation, where the computer employs trial and error to come up with a solution to the problem. Basically, during the whole learning process, the agent gets either rewards or penalties for the actions it performs. The overall goal is to maximize the total rewards.




We have used a model-free reinforcement learning algorithm known as Q-learning \cite{bib16}. It is an off-policy algorithm to determine the best action in the current state. Off-policy means that an agent, rather than following certain rules of behavior can take random actions; best action assumes that the action will result in the highest reward; current state is the present situation the agent resides. Basically there exists a system of rewards to build a matrix of scores for each possible move known as Q-matrix.


What Q-learning does is measure how good a state-action combination is in terms of rewards. It does so by keeping track of a Q-matrix a reference matrix, which gets updated after each episode with its row corresponding to the state and its column to the action. An episode ends after a set of actions is completed. Q-matrix is updated using a mathematical formula, known as the Bellman equation.


\[
\underbrace{\text{New }Q(s,a)}_{\substack{\text{New}\\
                                          \text{Q-Value}}}
    = \underbrace{Q(s,a)}_{\substack{\text{Current}\\
                                     \text{Q-Value}}}
    + \tikzmarknode{A}{\alpha}
    \Bigl[
        \underbrace{R(s,a)}_{\text{Reward}}
    + \tikzmarknode{B}{\gamma}
        \overbrace{\max Q'(s',a')}^{\mathclap{%
            \substack{\text{Maximum predicted reward, given} \\
                      \text{new state and all possible actions}}
                                            }}
    - Q(s,a)
    \Bigr]
\begin{tikzpicture}[overlay, remember picture,shorten <=1mm, font=\footnotesize, align=center]
\draw (A.south) -- ++ (0,-.8) node (C) [below] {Learning\\ rate};
\draw (B.south) -- (B |- C.north)  node[below] {Discount\\ rate};
\end{tikzpicture}
\vspace{4ex}
\]

In the above equation the first term, Q(s, a) is the value of the current action in the current state, alpha is the learning rate, that controls how much the difference between previous and new Q-value is considered. Gamma is a discount factor, which is used to balance between immediate and future reward. The updates occur after each step or action and ends when an episode is done (reaching the terminal point). The agent will not learn much after a single episode, but eventually with enough exploring (steps and episodes) it will converge and learn the optimal Q-values. 

RL has had extensive success in complex control environments like Atari games\cite{bib4}, Sokoban planning \cite{bib5}. It is also applied to games where there is real time strategy (RTS) such as bots \cite{bib9}, another reinforcement learning based approach \cite{bib10} chooses from a set of predefined strategies in turn based strategy based games. In such approaches the training process is separated into several stages, each of them responsible for different aspects of the game (such as combat, movement and exploration).  Other works in strategic fighting games \cite{bib12,bib13} map the possible states of the game based on low-level formations, such as distance between the fighters and health points. The reward function used are simple: a positive reward is granted every time the agent strikes the opponent and a negative reward is given when the agents gets hit. A very recent study \cite{bib19} introduced  natural language processing into the study of knot theory, and they also utilize reinforcement learning (RL) to find sequences of  moves and braid relations that simplify knots and can identify unknots by explicitly giving the sequence of actions. Another study \cite{bib20} proposed HULK a perception-based system that untangles dense overhand and figure-eight knots in linear deformable objects from RGB observations. It exploits geometry at local and global scales and learns to model only task-specific features, instead of performing full state estimation, to enable fine-grained manipulation.



\section{OpenAI Gym}
Recent advances in RL combines Deep Learning (DL) with RL (Deep Reinforcement Learning)  and have shown that model-free optimization, or policy gradients, can be used for complex environments \cite{bib16}. However, in order to continue testing new ideas and increasing the quality of results, the research community needs good benchmark platforms. This is the main goal of OpenAI Gym platform \cite{bib1}. It is basically a toolkit used for developing and testing reinforcement learning algorithms. One of the encouraging aspect of choosing OpenAI gym it makes no assumptions about the structure of the agent, and has compatibility with any numerical computation library, such as Theano or Google’s Tensorflow. Gym is a library, which contains a collections of test problems, known as environments, which can be used for testing reinforcement learning algorithms. It also leverages the user to design their own customized environments. A commonality in all of reinforcement learning is an agent situated in an environment. In each step, the agent takes an action and as a result receives an observation and a reward from the environment. What makes OpenAI Gym unique is how it focuses on the episodic setting of reinforcement learning, where the agent’s action chains are broken down into a sequence of episodes. Each episode begins by randomly sampling the agent’s initial state and continues until the environment reaches a terminal state. The purpose of structuring reinforcement learning into episodes like these is to maximize the expected total reward per episode, and to manage a high level of performance in as few episodes as possible.

\subsection{Environment Set-up for our problem}

To use the Q-Learning algorithm, it is necessary setup the environment which defines all the the possible actions and states of the agent. These states must encode useful information to the learning process. In our case of braids with two strands the following states are observed: (aa,bb,ab,ba,a1,1a,1b,b1,11). The agent remains in the same state until a legal action takes place. All the legal actions are described in the \emph{Introduction} Section of the paper. For braids with $2$ and $3$ strands basically we have a caret which moves back and forth over the string. Each time it moves over the string agent would be in specific state, and that state would only be changed after some legal action takes place. For the case of braids with $2$ strands, the caret moves over two characters at a time, whereas for the case with $3$ strands caret moves over three characters at a time in the whole string so the state-space is also large.

\subsection{Action Space and Rewards}
The table~\ref{tab:my-table} shows the rewards associated with each action for braids with $2$ and $3$ strands. As we have already discussed all such actions that bring us closer to the target output value will have the higher rewards and all such actions which takes us away from the target output will have lesser rewards.
\begin{table}[]
\centering
\begin{tabular}{|l|l|}
\hline
\textbf{Action} & \textbf{Reward} \\ \hline
CARET\_MOVE     & 0               \\ \hline
ROTATE\_TRUE    & 0               \\ \hline
ROTATE\_FALSE   & 0               \\ \hline
REPLACE\_TRUE   & 1               \\ \hline
REPLACE\_BACK   & -2              \\ \hline
REPLACE\_FALSE  & -1              \\ \hline
ROTATE\_REPLACE   & 1              \\ \hline
\end{tabular}
\caption{Reward associated for each action}
\label{tab:my-table}
\end{table}
 For the case of braids with $2$ strands. There are certain actions such as $action\_replace$, replaces ($ab$ to $11$, $ba$ to $11$), $action\_replace\_back$, replaces ($11$ to $ab$, $11$ to $ba$). Whereas, $action\_rotate$ moves the position of string e.g., ($1a$ to $a1$, $1b$ to $b1$) and vice versa. $Action\_move\_caret(left/right)$, this action moves caret to the left or right. The reward associated with $action\_replace$, true is 1, $action\_replace$ false is $-1$, reward for $action\_replace\_back$ is $-2$, $action\_rotate\_replace$ is $1$, for all other actions reward is $0$. 
Similarly for the other case where we have braids with 3 strands, $action\_replace$, replaces ($Aa$ to $11$, $aA$ to $11$, $Bb$ to $11$, $bB$ to $11$), $action\_replace\_back$ replaces($11$ to $Aa$, $11$ to $aA$, $11$ to $Bb$, $11$ to $bB$), $action\_rotate\_replace$ moves the position of the strings ($ABA$ to $BAB$, $BAB$ to $ABA$), $action\_rotate$ moves the position of the strings($aA$ to $Aa$, $aA$ to $Aa$, $bB$ to $Bb$, $Bb$ to $bB$). The choice of the reward selection is inspired from few of the works recently published \cite{bib2,bib8}.

\section{Experiments}

\begin{figure}[ht]
\centering
\begin{tabular}{c@{\quad}c}
\includegraphics[height=1.2 in,scale=0.3]{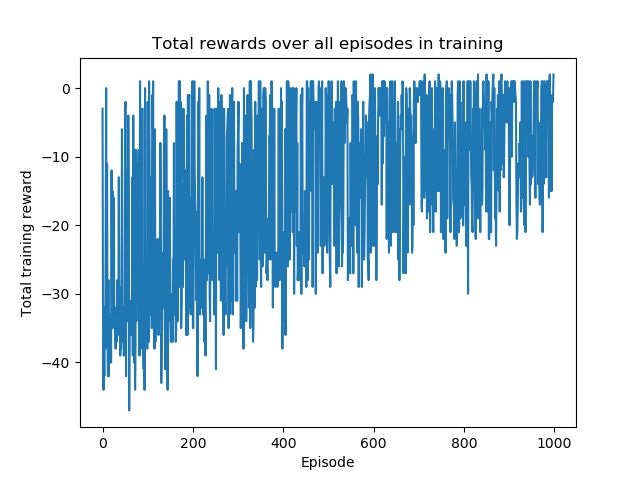}     
 & 
\includegraphics[height=1.2 in,scale=0.3]{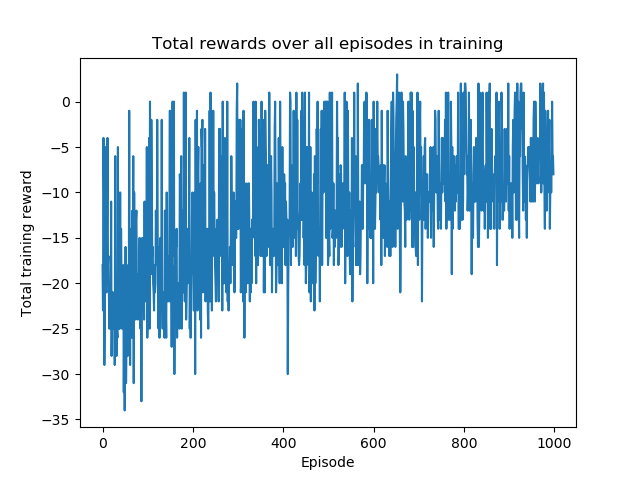}\\[.1ex]
Ep vs Rw @1000 episodes
 &
Ep vs Rw @1000 episodes
\end{tabular}
\caption{Plots for Rewards during training over episodes for n=7 and n=8}
\label{fig:obj-detection1}
\end{figure}

\begin{figure}[ht]
\centering
\begin{tabular}{c@{\quad}c}
\includegraphics[height=1.2 in,scale=0.3]{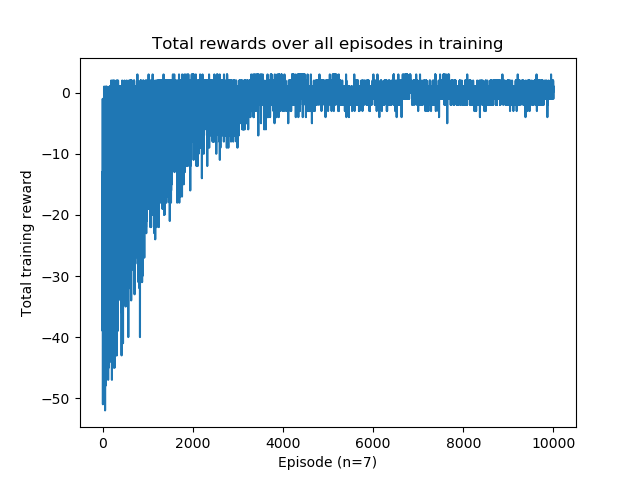}     
 & 
\includegraphics[height=1.2 in,scale=0.3]{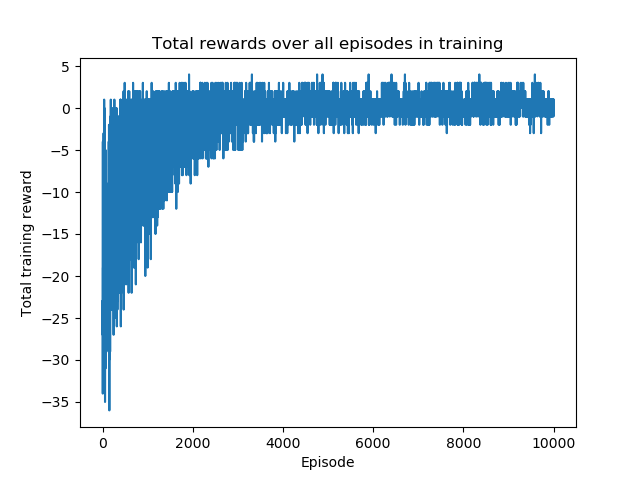}\\[.1ex]
Ep vs Rw @1000 episodes
 &
Ep vs Rw @10000 episodes
\end{tabular}
\caption{Plots for Rewards during training over episodes for n=7 and n=8, Ep=Episodes, Rw=Rewards}
\label{fig:obj-detection3}
\end{figure}

Untangling of braids requires the implementation of Q-learning algorithm discussed in Section 2. To measure the performance of Q-learning model, we utilize the metrics provided by OpenAI Gym interface, namely {\it rewards over episodes} of a particular environment. Separate experiments were performed for different environments. The choice of hyper-parameters selection was looked from some of the work in the literature \cite{bib3}. In the environment where we consider braids with $2$ strands, we have a single agent which performs series of actions during training to untangle the braid. We observe during the training process inside each episode an agent starts with random actions to untangle the braids, and finally over the period of time learns the right actions to reach the target output. It can be observed looking at figures ~\ref{fig:obj-detection1},~\ref{fig:obj-detection3} of different lengths of the input that negative rewards are quite prominent, if we train the model for lesser number of episodes, and agent hardly learns, whereas the episodes rewards progressively increase over time and ultimately levels out at a high reward per episode value from episode 4000, which indicates that the agent learns to maximize its total reward earned over the period of time.


In the multi-agent scenario, where we consider braids with 3 strands, in each episode the first agent for the given length of the input tries to tangle the braid during the fixed number of defined steps applying the transformations discussed in Section 1, that tangled state is the input for the second agent which again applies the same transformations to un-tangle the braid. As we approach the problem as a competitive game between two players (player1 = tangling player, player2= un-tangling). It is observed from Table ~\ref{tab:my-tabl}, for lesser number of training episodes and larger length of the input the probability of the tangling player to win the game is more, whereas when we train the system for higher number of episodes the probability of the un-tangling player to win the game is more times at the end of training. Figures \ref{braid1}, \ref{braid2} shows the examples hard tangled braids produced by player1 after 150 episodes. \\


\begin{table}[]
\centering
\begin{tabular}{|l|l|l|l|l|}
\hline
\textbf{Input length} & \textbf{\begin{tabular}[c]{@{}l@{}}ep=1000\\ steps=20\end{tabular}} & \textbf{\begin{tabular}[c]{@{}l@{}}ep=10000\\ steps=20\end{tabular}} & \textbf{\begin{tabular}[c]{@{}l@{}}ep=1000\\ steps=100\end{tabular}} & \textbf{\begin{tabular}[c]{@{}l@{}}ep=10000\\ steps=100\end{tabular}} \\ \hline
7                     & 40\%                                                                      & 81.7\%                                                                     & 46.2\%                                                                     & 66.6\%                                                                      \\ \hline
8                     & 30.7\%                                                                    & 85\%                                                                       & 48.6\%                                                                     & 85\%                                                                        \\ \hline
9                     & 29.4\%                                                                    & 87.2\%                                                                     & 42.7\%                                                                     & 75.8\%                                                                      \\ \hline
10                    & 24.9\%                                                                    & 72.9\%                                                                     & 36.3\%                                                                     & 84.9\%                                                                      \\ \hline
11                    & 24.8\%                                                                    & 72.3\%                                                                     & 32.7\%                                                                      & 60.9\%                                                                      \\ \hline
\end{tabular}
\caption{probability of player2 of winning the game, ep=episodes }
\label{tab:my-tabl}
\end{table}

\section{Conclusion}
In this pilot study we successfully conducted several experiments using Q-learning algorithm to untangle the braids with 2 and 3 strands. The problem of untangling of braids with 2 strands was simply approached as rule-based approach, where the agent learns over the time right rules to untangle the braid. Whereas, the problem to untangle the braids with 3 strands was approached as a competitive game between two players, where the first agent starts with a fixed length of input and applies the rules to tangle the braid, that tangled braid is the input for the second agent which again applies the rules to untangle the braid, ultimately if second agent successfully untangles the braid it wins the round. We observe the more we train the model, the more is the probability of the second agent to win the game. In the future we intend to approach the similar problem using DQN(Deep Q-leaning Network) to compare the result with Q-learning approach.

\bibliographystyle{splncs04}
\bibliography{main}
\end{document}